# Defining and Categorising Human-AI Interactions in Clinical Trials: A Multidimensional Human-AI Classification Approach

Sandra Woolley[1,*], Tim Collins[2], Khalid Khattak[1], Illia Chernomorets[1], Ariane Arevalo[1] and Chris Richardson[1]

[1]Keele University, UK
[2]Manchester Metropolitan University, UK

**Abstract**

This paper examines human-AI interactions (HAIIs) in clinical trials and presents a multidimensional categorisation framework that classifies interactions according to AI tasks, human-AI relationships, interaction configurations and interacting human groups. We define HAII, examine existing taxonomies and extend existing categorisation approaches through this novel multidimensional framework. We purposively sampled 15 clinical trials from a previously reported dataset. Each trial was independently categorised by two human reviewers and six large language model (LLM) classifiers. The proposed categorisation provides a structured method for the consistent identification, comparison and synthesis of human-AI interactions across clinical-trial records. The framework is intended to support more consistent comparison and synthesis of AI-related clinical trials and to make explicit the different forms of human involvement associated with AI interventions. Across the selected trials, the combined human and LLM consensus most frequently categorised AI as a tool (15/15), identified AI-first human-in-the-loop as the most frequent interaction configuration (13/15), and identified classification as the most frequent AI task (11/15). Health professionals were the human group most frequently identified as interacting with AI. The findings indicate that interaction sequence and human oversight are difficult to determine from clinical trial records. Agreement was lowest for human-on-the-loop and AI-follows human-in-the-loop interactions. Despite being prompted to assign uncertain classifications where the available information was insufficient, LLM classifiers were less likely than human reviewers to do so. These results demonstrate the potential for LLM-assisted categorisation while indicating the continuing importance of human judgement where trial records are incomplete or ambiguous. The principal contribution is a proposed multidimensional framework that brings together AI tasks, human-AI relationships, interaction configurations and interacting human groups within a single approach designed for clinical-trial records. Its significance lies in its potential to support more systematic identification, comparison and synthesis of how humans and AI interact in clinical trials.

## 1. Introduction

Artificial intelligence (AI) is being investigated for applications in clinical research and healthcare. Evaluating AI in clinical settings requires consideration not only of technical performance but also of its integration into clinical processes and how humans engage with its outputs, participate in decisions and oversee its actions [1, 2]. Clinical trial registries provide an important source of evidence about the prospective evaluation of AI. Established in 2000, ClinicalTrials.gov currently contains almost 600,000 registered studies across 226 countries and territories [3]. However, AI-related trials can be difficult to identify, compare and synthesise because registry records and published reports vary in how they describe AI methods, human involvement, inputs and outputs, and the clinical context in which AI is used [4]. The SPIRIT-AI and CONSORT-AI reporting guidelines were developed to improve the reporting of clinical trials involving AI. SPIRIT-AI extended the Standard Protocol Items: Recommendations for Interventional Trials (SPIRIT) guidance for clinical-trial *protocols*, while CONSORT-AI extended the Consolidated Standards of Reporting Trials (CONSORT) guidance for *reports* of completed randomised controlled trials (RCTs) [5, 6]. The recommendations specify that trial

*Corresponding author.
s.i.woolley@keele.ac.uk (S. Woolley)
0000-0002-7623-2866 (S. Woolley); 0000-0003-2841-1947 (T. Collins); 0000-0001-8410-5915 (K. Khattak); 0009-0004-8871-4012 (I. Chernomorets); 0009-0002-8068-7529 (A. Arevalo); 0009-0000-0555-8626 (C. Richardson)

reports should describe AI interventions, their intended use and setting, the handling of inputs and outputs, the expertise required by users, and the interaction between humans and AI. Nevertheless, assessments of published RCTs indicate that reporting remains incomplete. For example, a review of ophthalmology trials reported that no trials reported all CONSORT-AI items, while a subsequent review of oncology trials identified persistent reporting omissions concerning algorithm versions, input data, the handling of poor-quality data and the assessment of errors [7, 8].

Reporting limitations are not limited to AI. For example, in studies and trials of wearable technologies, substantial omissions in technology and device reporting information have also been identified [9, 10].

Many trials predate the SPIRIT-AI protocol guidance and CONSORT-AI RCT report guidance. It is also notable that RCTs comprise only a minority of clinical studies. For example, Lam *et al.* [11] reported that amongst 11,839 articles on AI in clinical practice, only 39 (0.33%) included RCTs. Similarly, Wang *et al.* [12] reported that, between September 2023 and September 2024, only 2.4% of AI clinical studies were RCTs.

Although reporting guidance supports fuller descriptions of individual trials, it does not provide a comprehensive framework for comparing the different ways in which humans and AI interact. Consequently, even when AI functionality and use can be reasonably identified from trial records, researchers have limited means to systematically compare patterns of human-AI interaction across trials, clinical contexts and intervention types. These interactions may differ in terms of the tasks performed, the order of human and AI contributions, the configuration of exchanges, the types of humans involved and the extent of human oversight. A structured categorisation of human-AI interactions is therefore needed to support their consistent identification, comparison and synthesis across clinical trials, and to make explicit the different roles allocated to humans and AI within clinical processes and decision-making activities.

## 2. Related Work

Plana *et al.* [13] systematically reviewed randomised clinical trials evaluating machine-learning interventions used in healthcare. Despite the large number of machine-learning systems being developed for healthcare, the review identified only 41 published RCTs, although the authors reported a rapidly increasing trend. None of the 41 trials fully complied with CONSORT-AI guidance, with common omissions including failure to analyse performance errors and report the availability of algorithms. The authors also noted that only 27% of trials reported race and ethnicity, indicating limited capacity to assess demographic representativeness. However, current reporting frameworks are limited and not universally meaningful [14]. For example, the US Office of Management and Budget (OMB), defines the minimum federal race and ethnicity reporting categories as: American Indian or Alaska Native; Asian; Black or African American; Hispanic or Latino; Middle Eastern or North African; Native Hawaiian or Pacific Islander; and White. These are used extensively in US federal reporting, NIH-funded clinical research and other studies, but while meaningful in a US context, these categories lack international meaning and relevance.

AI has extensive potential to support medical research, healthcare and clinical trials. In clinical trials it can, for example, support protocol design, participant recruitment, data analysis and outcome prediction [15]. AI can also support trial reporting, for example, through automated assessment against reporting standards or assistance with the generation of plain-language summaries [16, 17]. However, to enable the consistent reporting of HAII, whether by human authors or AI-supported tools, a framework is needed to describe HAII consistently and unambiguously across studies.

Alongside its use in supporting trial processes, AI is increasingly evaluated as an intervention within clinical practice. A scoping review found that most published RCTs of AI in clinical practice reported positive primary endpoints, but these endpoints predominantly concerned diagnostic performance or diagnostic yield rather than patient-relevant health outcomes [18].

Two recent reviews synthesise the academic literature on human-AI interaction in clinical and healthcare contexts [19, 20]. Gomez *et al.* [19] systematically reviewed 105 empirical studies of *AI-assisted decision-making* across healthcare and other domains. They observed that relatively simple forms of AI assistance predominated, and that reciprocal and adaptive forms of human-AI collaboration were less common. Their taxonomy principally characterises patterns of exchange between humans and AI, focusing on sequencing, initiative and the structure of interaction. Their taxonomy comprised the following categories:

- AI-first assistance: AI presents a prediction before the human makes the final decision.
- AI-follow assistance: the human forms an initial judgement before receiving AI assistance.
- Secondary assistance: AI provides supporting information rather than a direct solution.
- Request-driven AI assistance: the human actively requests AI assistance.
- AI-guided dialogic user engagement: AI guides a repeated exchange in which the human supplies information.
- User-guided interactive adjustments: human feedback, corrections or information shape the AI output.
- Delegation: responsibility for a decision is transferred to either the human or AI.
- Other patterns: complex arrangements involving multiple decisions, continuous interaction, additional agents or aggregated human and AI judgements.

The taxonomy principally characterises interaction processes rather than the identity of the human participants, the substantive role performed by the AI, or the clinical context within which interaction occurs.

You and Li [20] conducted a scoping review of the academic literature on collaboration and task allocation between doctors and AI in healthcare. The review included 85 publications and found that collaboration patterns were associated with the risk and degree of automation of clinical tasks. It also identified a clear division of labour between doctors and AI, while evaluations of collaboration extended beyond technical performance to include clinical efficiency and user experience. Their review categorises the literature according to:

- Human-AI collaboration patterns across different medical tasks.
- Task-division mechanisms between physicians and AI.
- Evaluation measures, including technical performance, clinical efficiency and user experience.

Together, these reviews address interaction patterns, task allocation and evaluation, but do not provide a multidimensional framework designed for categorising HAII *per se* in clinical-trial records.

For the onerous and time-consuming task of literature screening in systematic reviews, LLMs may provide support, although their evaluation requires attention to class imbalance, false-negative costs and the loss of relevant evidence [21]. In our prior hybrid human-AI categorisation of HAII in clinical trials [4], trials were categorised according to AI usage, and attempts were made to categorise the human groups interacting with AI. This exploration revealed areas of ambiguity and subjective interpretation, particularly where trial records provided limited information regarding AI functionality, interaction sequencing and human oversight. These observations motivated the development of the multidimensional framework presented in this study.

## 3. Study Aims and Research Questions

The motivation for the research underpinning this study was the ambition to analyse and understand HAII in clinical trials and to observe trends in its use.

### 3.1 Study Aims

The aim of this study was to build on prior work that categorised trials according to the human groups interacting with AI by developing and testing a multidimensional framework for categorising human-AI interactions in clinical trials and undertaking a preliminary examination of LLM-assisted categorisation.

In prior work, HAII categorisation focused on identifying AI use and the human actors involved:

- No use of AI
- No human-AI interaction
- Patient AI-interaction
- Caregiver-AI interaction (including paid childminders)
- Health professional-AI interaction
- Other Human-AI interaction (e.g. including medical students)
- Hybrid-AI interaction (more than one group, e.g. patients and health professionals)

However, this categorisation did not capture other dimensions of HAII, such as the task performed by the AI, the relationship between the human and AI, or the configuration of the interaction. The additional dimensions and categories were informed by existing taxonomies and previous research, and were refined through observation, pilot categorisation and application to a sample of trials.

### 3.2 Research Questions

The study addressed the following research questions concerning the definition, categorisation and classification of HAII in clinical-trial records.

**RQ1. What is human-AI interaction?**

a. How is human-AI interaction defined?

b. What constitutes a minimal unit of human-AI interaction?

**RQ2. Which dimensions are relevant to the classification of human-AI interaction in clinical trials?**

a. What are the categories for these dimensions?

b. How can the dimensions and categories be applied to the classification of clinical-trial records?

**RQ3. What preliminary observations can be made through comparison of human and LLM classifications of HAII in clinical-trial records?**

## 4. What is Human-AI Interaction?

This section defines HAII for the purposes of the framework and considers the minimum conditions under which an interaction between a human and an AI system can be said to occur.

### 4.1. Definitions of Human-AI Interaction

HCI is a mature field concerned with the design, use and evaluation of interactive computational systems. Dix *et al.* [22] define HCI as: *"The study of people, computer technology and the ways these influence each other"*.

Dictionary definitions commonly characterise interaction itself as mutual or reciprocal, however, this reciprocal, two-way definition of interaction is restrictive and excludes important instances of one-way human-to-AI and AI-to-human influence. For example, if a clinician is automatically presented with an AI assessment of a patient and this influences their judgement or action, this may reasonably be classified as HAII, even if the clinician provides no input to the AI system. For the purposes of this work, human-AI interaction is therefore defined as occurring where an AI system or its output influences human perception, interpretation, judgement or action, or where human input influences the operation or output of an AI system.

Although human-computer interaction (HCI) provides a well-established foundation for understanding interactions between humans and computational systems, AI systems introduce characteristics that are less prominent in conventional software, including adaptation, prediction, probabilistic reasoning and generative capabilities. Consequently, conventional HCI concepts may not always describe interactions involving AI systems with sufficient precision. The definition adopted may affect whether human involvement is recognised and how oversight and responsibility are understood.

A particular challenge concerns the minimum conditions necessary for human-AI interaction to occur. Some accounts implicitly require observable action or behavioural change, whereas others may regard the perception, interpretation or processing of information as sufficient. This raises the question of what constitutes a minimal unit of HAII, which is considered in the following section.

### 4.2. What is a minimal unit of Human-AI Interaction?

Establishing what constitutes a minimal HAII requires consideration of when AI output enters human experience and when human input enters AI processing. Table 1 examines these lower boundaries from both AI-to-human and human-to-AI perspectives.

**Table 1**: Defining the lower boundaries of Human-AI Interaction (HAII)

| AI-to-human perspective | HAII | Reason |
|---|---|---|
| AI generates an output, but no human encounters it | No (not ordinarily) | There is an AI output, but no realised human encounter with it. |
| AI output is displayed and a human perceives it | Yes (in a minimal sense) | The AI-generated output has entered the person's experience. |
| Human consciously rejects the output | Yes | Rejection is a response. |
| Human deliberately ignores or dismisses it | Yes | Deliberate non-engagement is still a response to something perceived. |
| Human fails to notice the output | No (probably not) | The output was presented, but no interaction was realised from the human side. |
| Human notices it but takes no visible action | Yes | Interaction may affect attention, interpretation or judgement without observable behaviour. |
| **Human-to-AI perspective** | **HAII** | **Reason** |
| A human generates an input, but no AI encounters it | No (not ordinarily) | There is human-generated input, but no realised AI encounter with it. |
| Human-generated input is presented to an AI system that detects it | Yes (in a minimal) sense | The human-generated input has entered the AI's detection or receptive process. |
| AI rejects human-generated input | Yes | Rejection is a computational response to input that the AI has detected. |
| AI deliberately ignores or dismisses human-generated input | Yes | Functional non-engagement is still a response if the AI first detects the input and then excludes it according to its rules or operating parameters. |
| AI fails to detect human-generated input | No (probably not) | The input was presented, but it was not detected or registered by the AI. |
| AI detects human-generated input but takes no externally observable action | Yes | Interaction may affect the AI's internal state, classification or processing without producing a visible output. |

For the purposes of this framework, therefore, human perception of an AI output, or AI detection of human input, constitutes a minimal HAII, even where no externally observable response follows.

## 5. Methodology

The methodology comprised two stages. First, the dimensions and categories of the HAII framework were developed and refined through examination of existing taxonomies, previous research and pilot categorisation of clinical-trial records. Second, the framework was applied by two human reviewers and six LLM classifiers to a purposively selected sample of 15 diverse HAII trials drawn from a previously reported dataset of 100 trials (Woolley et al., 2026). Classifications were based on the Study Details and Research View information available in the ClinicalTrials.gov record for each trial. This included key trial information such as the study overview, trial locations, interventions, outcome measures and trial descriptions. Attached documents and linked publications were not included.

## 5.1 Defining HAII Dimensions and Categories

Following examination of the literature and multiple rounds of pilot categorisation of clinical-trial records, we identified the following dimensions and categories as summarised in Table 2. Importantly, categories within each dimension are not mutually exclusive, and multiple categories can be assigned to a given trial record.

**Table 2**: HAII dimensions and categories

| Dimension | Category |
|---|---|
| AI task | Classify |
| | Detect/assess |
| | Diagnose |
| | Educate |
| | Explain/interpret |
| | Monitor/respond |
| | Natural language communication |
| | Personalise |
| | Predict/prognose |
| | Recommend/advise |
| | Score/rank |
| | Summarise/synthesise |
| | Treat |
| Human-AI relationship | AI as a collaborator |
| | AI as a decision-support resource |
| | AI as a therapeutic component |
| | AI as a tool |
| | AI as an educator/communicator |
| | AI as proactive or autonomous |
| Human-AI interaction | AI outputs, human decides (AI first; human-in-the-loop) |
| | Autonomous AI action without human oversight (human-out-of-the-loop) |
| | Human reviews/decides, then considers AI output (AI follows; human-in-the-loop) |
| | Humans oversee AI action (human-on-the loop) |
| | Integrated interactions (multiple humans and/or multiple AIs) |
| | Iterative interaction (repeated exchanges, including chatbot interaction) |
| Humans interacting with AI | Caregiver-AI interaction |
| | Health professional-AI interaction |
| | Other human-AI interaction |
| | Patient-AI interaction |

## 5.2 Selected Trials and Selected LLMs

A set of 15 trials listed in Table 3 was purposively selected to include variation in AI type, clinical application, human groups and HAII configuration. For the "Type of AI" field, consensus was based solely on explicit descriptions contained within the trial record. Where no specific AI methodology was stated, the consensus classification remained "unspecified" and no inference regarding the underlying AI approach was made. As evident from Table 3, the type of AI itself was often unspecified. The sample included trials involving different AI types and human groups, including trials expected to present classification challenges owing to limited reporting detail, ambiguity regarding AI functionality, or uncertainty concerning interaction sequencing and oversight.

Each selected trial was independently categorised against the dimensions in Table 2 by two expert human reviewers and six LLMs listed in Table 4. Classifiers were instructed to assign **True** (**T**), **False** (**F**) or **Uncertain** (?) for each category and to provide a confidence rating for categorisations in each dimension (low, medium, high). Further information supplied to the human and LLM classifiers is detailed in the prompt provided in Appendix A.

The six LLMs (Table 4) were purposefully chosen. Given the constrained nature of the classification task, we focused on five small, locally hosted models (L1–L5), avoiding the unnecessary overhead of large cloud-hosted systems. A single cloud-hosted frontier model (L6) was also included for comparison.

**Table 3**: The selected set of clinical trials

| NCT Number | Brief Title | Type of AI (consensus) |
|---|---|---|
| NCT00647413 | Prevent Exposure to Tobacco Smoke at Home | Expert System |
| NCT03780582 | Evaluation of Use of Diagnostic AI for Lung Cancer in Practice | Unspecified |
| NCT04238065 | A Clinical Trial of Caterna Virtual Reality Facilitating Treatment in Children with Amblyopia | Unspecified |
| NCT04269304 | Deciphering AMD by Deep Phenotyping and Machine Learning- Prospective Study – PINNACLE | Unspecified "Machine Learning" |
| NCT04289025 | Modelling and AI Using Sensor Data to Personalise REHABilitation Following Joint Replacement | Unspecified |
| NCT04679181 | Acceptability Assessment of an "Organization of Care Integrating Artificial Intelligence and a Solution of Telemedicine" on Care of the Nursing Home Residents Located in a Medical Desert | Unspecified |
| NCT04763980 | Community-Based Health Coach for Access to Germline Genetic Testing Among African American Men with Prostate Cancer | Chatbot |
| NCT05308303 | AI to Improve Data from Danish Cardiac Arrest Registry | Bag-of-words |
| NCT05384002 | An AI Platform Integrating Imaging Data and Models, Supporting Precision Care Through Prostate Cancer's Continuum | Unspecified |
| NCT06070415 | Adherence to Exercises for Low Back Pain | Chatbot |
| NCT06186557 | Automated Detection of Patient Ventilator Asynchrony Using Pes Signal | Convolutional Neural Network |
| NCT06265909 | Prospective Real-World Study on Therapy Prediction Algorithm Training | Deep Neural Network |
| NCT06423040 | Social Media Chatbot on Physical Activity Education for Older Adults | Chatbot |
| NCT06614530 | Effectiveness of Chilipad in Enhancing Sleep Quality and Wellbeing | Unspecified |
| NCT06664112 | Delivery Outcomes by AIDA (Artificial Intelligence Dystocia Algorithm) Analysis | Unspecified "Machine Learning" |

**Table 4:** LLM classifiers employed in the study

| LLM ID | Model | Total / Active Parameters | Host |
|---|---|---|---|
| L1 | liquid/lfm2-24b-a2b | 24B / 2.3B | Local |
| L2 | openai/gpt-oss-20b | 20B / 3.6B | Local |
| L3 | mistralai/mistral-small-3.2 | 24B / 24B | Local |
| L4 | microsoft/phi-4 | 15B / 15B | Local |
| L5 | mistralai/mistral-nemo-instruct-2407 | 12B / 12B | Local |
| L6 | GPT-5.6 Sol | * | Cloud |

*GPT-5.6 Sol is a proprietary model; parameters are estimated as approximately 2.2T (total) 150B (active)

# 6. Results

The eight classifiers (six LLMs and two humans) produced 3,480 category-level classifications across the 15 trials, as summarised in Table 5. Figure 1(i)–(iii) presents the category-level classifications for trials 1–5, 6–10 and 11–15, respectively.

**Table 5:** Number of classifications performed by LLM and human classifiers

| Classifier Type | No. of Classifiers | No. of Trials | No. of Categories | Total Classifications |
|---|---|---|---|---|
| LLMs (L1-L6) | 6 | 15 | 29 | 2,610 |
| Humans (H1, H2) | 2 | 15 | 29 | 870 |
| **Total** | **8** | **15** | **29** | **3,480** |

These results are reported as individual classifier responses, consensus classifications and agreement among the eight classifiers.

## 6.1 Classification Frequencies

Table 6 identifies the categories most frequently classified as True, False and Uncertain (‘?’) across all eight classifiers. Out of the total of 120 classifications (8 classifiers x 15 trials = 120 classifications), AI as a tool was most frequently classified as True (114/120 classifications), followed by classify (82/120), AI-first human-in-the-loop interaction (76/120), health professional-AI interaction (75/120) and AI as a decision-support resource (73/120). Treat was most frequently classified as False (108/120), while AI-follows human-in-the-loop interaction was most frequently classified as Uncertain (47/120).

The categories most frequently classified as True by consensus among the eight classifiers were AI as a tool (15/15 trials), AI-first human-in-the-loop interaction (13/15), classification (11/15), AI as a decision-support resource (11/15), health professional-AI interaction (10/15), detection or assessment (9/15), and human-on-the-loop interaction (9/15).

No categories were entirely unused by the individual classifiers: each category received at least one True classification somewhere in the dataset. However, three categories were not classified as True by consensus for any of the 15 trials:

- Treat as the AI task
- AI as a collaborator
- Other human-AI interaction

## 6.2 Classifier Agreement

Table 7 presents the categories with the highest and lowest mean agreement among the eight classifiers. Agreement was highest for AI as a tool and predict/prognose, both with a mean agreement of 7.6/8. Agreement was lowest for human-on-the-loop interaction (4.1/8), followed by AI-follows human-in-the-loop interaction (4.9/8). Three of the five categories with the lowest agreement concerned the configuration of human-AI interaction.

## 6.3 Human and LLM Classifications

Human reviewers were more likely to assign Uncertain (‘?’) classifications than LLMs. The greatest disagreement between the LLM and human consensus classifications occurred in the human-AI interaction dimension. Opposing True and False consensus classifications were most frequent for AI-first human-in-the-loop interaction (6/15 trials), followed by human-on-the-loop (4/15) and human-out-of-the-loop interaction (2/15).

| | | AI Task Category | | | | | | | | | | | | | | Human-AI Relationship | | | | | | | Human-AI Interaction | | | | | | | Humans Interacting with AI | | | | |
|---|---|---|---|---|---|---|---|---|---|---|---|---|---|---|---|---|---|---|---|---|---|---|---|---|---|---|---|---|---|---|---|---|---|---|
| NCT Number | Classifier | Classify | Detect/assess | Diagnose | Educate | Explain/interpret | Monitor/respond | Natural language communication | Personalise | Predict/prognose | Recommend/advise | Score/rank | Summarise/synthesise | Treat | AI Task Category Confidence | AI as a collaborator | AI as a decision-support resource | AI as a therapeutic component | AI as a tool | AI as an educator/communicator | AI as proactive or autonomous | Human-AI Relationship Confidence | AI first: human-in-the-loop | Human-out-of-the-loop | AI follows: human-in-the-loop | Human-on-the loop | Multiple humans and/or AIs | Iterative interaction | Human-AI Interaction Confidence | Caregiver-AI Interaction | Health Professional-AI Interaction | Other Human-AI Interaction | Patient AI-Interaction | Humans Interacting with AI Confidence |
| NCT00647413 | L1 | T | F | F | T | T | F | T | T | F | T | F | F | F | H | F | T | F | T | T | F | H | T | F | ? | T | F | F | M | T | F | T | F | H |
| | L2 | T | F | F | F | T | F | F | T | F | T | F | F | F | H | F | ? | F | T | F | F | M | T | F | F | F | F | F | H | F | T | F | T | H |
| | L3 | T | T | F | ? | T | F | ? | T | F | T | F | F | F | M | F | T | ? | T | ? | F | M | T | F | ? | T | ? | F | M | T | ? | F | F | M |
| | L4 | F | T | F | T | T | F | F | T | F | F | F | F | F | M | F | T | F | T | T | F | M | T | F | F | F | F | F | M | T | F | F | F | M |
| | L5 | T | ? | F | T | F | F | F | F | F | ? | F | F | F | L | F | ? | T | F | T | F | L | T | F | ? | ? | ? | F | M | T | ? | F | F | L |
| | L6 | ? | ? | F | T | T | F | T | T | F | T | ? | ? | F | M | F | F | T | T | T | ? | M | ? | ? | F | ? | F | F | L | T | ? | F | ? | M |
| | H1 | T | F | F | T | T | F | T | T | F | T | F | F | F | M | F | T | F | T | T | F | M | F | ? | F | ? | F | F | M | F | ? | F | T | M |
| | H2 | ? | ? | F | ? | T | F | T | T | F | T | ? | ? | F | L | F | F | F | T | T | F | H | ? | ? | F | ? | F | F | L | T | F | F | ? | H |
| Consen. | L | 3 | 0 | -6 | 3 | 4 | -6 | -1 | 4 | -6 | 3 | -5 | -5 | -6 | M | -6 | 2 | -1 | 4 | 3 | -5 | M | 5 | -5 | -3 | 0 | -4 | -6 | M | 4 | -1 | -4 | -3 | M |
| | H | 1 | -1 | -2 | 1 | 2 | -2 | 2 | 2 | -2 | 2 | -1 | -1 | -2 | LM | -2 | 0 | -2 | 2 | 2 | -2 | MH | -1 | 0 | -2 | 0 | -2 | -2 | LM | 0 | -1 | -2 | 1 | MH |
| | L+H | 4 | -1 | -8 | 4 | 6 | -8 | 1 | 6 | -8 | 5 | -6 | -6 | -8 | M | -8 | 2 | -3 | 6 | 5 | -7 | M | 4 | -5 | -5 | 0 | -6 | -8 | M | 4 | -2 | -6 | -2 | MH |
| NCT03780582 | L1 | T | ? | F | F | F | F | F | F | F | F | F | F | F | M | F | T | F | T | F | F | H | ? | F | T | T | F | T | M | F | T | F | F | H |
| | L2 | T | T | T | F | F | F | F | F | F | F | T | F | F | H | F | T | F | T | F | F | H | T | F | F | F | F | F | H | F | T | F | F | H |
| | L3 | T | ? | F | F | F | F | F | F | T | ? | T | F | F | H | ? | T | F | T | F | F | H | T | F | ? | T | F | F | H | F | T | F | F | H |
| | L4 | T | T | F | F | T | F | F | F | T | F | T | F | F | H | T | T | F | T | F | F | M | T | F | F | F | F | F | M | F | T | F | F | H |
| | L5 | T | ? | F | ? | F | F | F | F | F | T | T | F | F | L | ? | T | F | T | ? | F | L | T | F | ? | ? | ? | F | L | F | T | F | F | L |
| | L6 | T | T | T | F | F | F | F | F | F | F | T | F | F | H | F | T | F | T | F | F | H | T | F | T | F | F | F | H | F | T | F | F | H |
| | H1 | T | T | F | F | F | F | F | F | F | F | T | F | F | H | F | T | F | T | F | F | M | T | F | T | F | F | F | M | F | T | F | F | H |
| | H2 | T | T | T | F | F | F | F | F | F | T | T | T | F | H | F | T | F | T | F | F | H | T | F | T | F | F | F | H | F | T | F | ? | H |
| Consen. | L | 6 | 3 | -2 | -5 | -4 | -6 | -6 | -6 | -2 | -3 | 4 | -6 | -6 | MH | -2 | 6 | -6 | 6 | -5 | -6 | MH | 5 | -6 | 0 | -1 | -5 | -4 | MH | -6 | 6 | -6 | -6 | MH |
| | H | 2 | 2 | 0 | -2 | -2 | -2 | -2 | -2 | -2 | 0 | 2 | 0 | -2 | H | -2 | 2 | -2 | 2 | -2 | -2 | MH | 2 | -2 | 2 | -2 | -2 | -2 | MH | -2 | 2 | -2 | -1 | H |
| | L+H | 8 | 5 | -2 | -7 | -6 | -8 | -8 | -8 | -4 | -3 | 6 | -6 | -8 | MH | -4 | 8 | -8 | 8 | -7 | -8 | MH | 7 | -8 | 2 | -3 | -7 | -6 | MH | -8 | 8 | -8 | -7 | H |
| NCT04238065 | L1 | T | F | F | F | F | F | T | F | F | F | F | F | F | H | F | T | T | T | T | F | H | ? | F | ? | T | F | F | M | T | T | F | F | H |
| | L2 | T | F | F | F | F | F | F | F | F | F | T | F | F | H | F | F | F | T | F | F | H | T | F | F | F | F | F | H | F | F | F | T | H |
| | L3 | ? | T | F | F | F | F | F | ? | F | F | T | F | F | M | F | ? | T | T | F | F | M | ? | F | ? | T | F | F | M | ? | T | F | T | M |
| | L4 | T | F | F | T | F | F | T | T | F | F | T | F | F | M | F | F | F | T | T | F | M | T | F | F | T | F | T | M | F | T | F | T | M |
| | L5 | T | ? | F | F | F | F | F | F | F | ? | F | F | T | M | F | ? | T | T | F | F | M | ? | F | T | ? | ? | F | L | ? | T | F | T | L |
| | L6 | ? | T | F | F | ? | ? | ? | T | F | F | T | F | T | M | T | F | T | T | ? | ? | M | F | ? | F | ? | ? | T | M | ? | ? | ? | T | M |
| | H1 | F | ? | F | F | F | ? | F | T | F | F | F | F | T | M | F | F | T | T | F | T | H | F | T | F | ? | F | F | M | ? | ? | F | T | M |
| | H2 | ? | ? | F | F | ? | F | ? | ? | F | F | T | F | T | M | F | F | T | F | F | T | H | F | T | F | F | F | F | H | F | F | F | T | H |
| Consen. | L | 4 | -1 | -6 | -4 | -5 | -5 | -1 | -1 | -6 | -5 | 2 | -6 | -2 | MH | -4 | -2 | 2 | 6 | -1 | -5 | MH | 1 | -5 | -2 | 2 | -4 | -2 | M | -1 | 3 | -5 | 4 | M |
| | H | -1 | 0 | -2 | -2 | -1 | -1 | -1 | 1 | -2 | -2 | 0 | -2 | 2 | M | -2 | -2 | 2 | 0 | -2 | 2 | H | -2 | 2 | -2 | -1 | -2 | -2 | MH | -1 | -1 | -2 | 2 | MH |
| | L+H | 3 | -1 | -8 | -6 | -6 | -6 | -2 | 0 | -8 | -7 | 2 | -8 | 0 | MH | -6 | -4 | 4 | 6 | -3 | -3 | MH | -1 | -3 | -4 | 1 | -6 | -4 | M | -2 | 2 | -7 | 6 | MH |
| NCT04269304 | L1 | T | F | F | F | F | F | F | F | T | F | F | F | F | M | F | T | F | T | F | ? | M | F | F | ? | T | F | F | M | F | T | F | F | H |
| | L2 | T | T | F | F | F | F | F | F | T | F | T | F | F | H | F | T | F | T | F | F | M | T | F | ? | ? | F | F | M | F | T | F | F | H |
| | L3 | T | T | F | F | ? | F | F | F | T | F | F | ? | F | M | F | T | F | T | F | ? | M | T | ? | F | T | ? | F | M | F | T | ? | F | M |
| | L4 | T | T | ? | F | F | F | F | F | T | F | F | F | F | M | F | T | F | T | F | ? | M | T | ? | F | T | ? | F | M | F | T | F | F | M |
| | L5 | T | T | F | F | ? | F | F | F | T | F | F | F | F | M | ? | T | F | T | F | F | M | T | F | ? | ? | T | F | M | F | T | F | ? | M |
| | L6 | ? | ? | F | F | F | F | F | F | T | F | ? | F | F | M | F | ? | F | T | F | F | M | F | F | F | F | F | F | M | F | F | ? | F | M |
| | H1 | ? | ? | F | F | F | F | F | F | T | F | F | F | F | M | F | T | F | T | F | F | M | T | ? | F | ? | F | F | L | F | T | F | F | M |
| | H2 | F | T | F | F | F | F | F | F | T | F | F | ? | F | M | F | T | F | T | F | F | H | ? | ? | F | ? | F | F | L | F | T | F | F | H |
| Consen. | L | 5 | 3 | -5 | -6 | -4 | -6 | -6 | -6 | 6 | -6 | -3 | -5 | -6 | M | -5 | 5 | -6 | 6 | -6 | -3 | M | 2 | -4 | -3 | 2 | -2 | -6 | M | -6 | 4 | -4 | -5 | MH |
| | H | -1 | 1 | -2 | -2 | -2 | -2 | -2 | -2 | 2 | -2 | -2 | -1 | -2 | M | -2 | 2 | -2 | 2 | -2 | -2 | MH | 1 | 0 | -2 | 0 | -2 | -2 | L | -2 | 2 | -2 | -2 | MH |
| | L+H | 4 | 4 | -7 | -8 | -6 | -8 | -8 | -8 | 8 | -8 | -5 | -6 | -8 | M | -7 | 7 | -8 | 8 | -8 | -5 | M | 3 | -4 | -5 | 2 | -4 | -8 | LM | -8 | 6 | -6 | -7 | MH |
| NCT04289025 | L1 | T | T | F | F | T | F | F | T | F | F | F | F | F | H | F | T | F | T | T | F | H | F | F | ? | T | F | F | M | F | T | F | T | H |
| | L2 | T | T | F | F | F | F | F | T | F | F | T | F | F | H | F | F | F | T | F | F | H | T | F | F | F | F | F | H | F | T | F | T | H |
| | L3 | T | T | F | F | ? | F | F | T | F | T | T | ? | F | M | F | T | ? | T | F | F | M | T | F | ? | T | ? | F | M | ? | T | F | T | M |
| | L4 | T | T | F | F | T | F | F | T | F | T | T | F | F | M | F | T | F | T | F | F | M | T | F | F | T | F | F | M | T | T | F | T | M |
| | L5 | T | T | F | ? | ? | F | F | T | F | ? | ? | F | F | M | F | T | ? | T | ? | F | M | T | F | ? | F | ? | F | L | T | T | F | F | L |
| | L6 | T | T | F | T | T | F | ? | T | F | T | T | T | T | M | F | ? | T | T | T | F | M | ? | F | F | ? | ? | T | M | F | T | F | T | H |
| | H1 | T | T | F | F | ? | F | F | ? | F | ? | ? | F | F | L | F | T | F | T | F | F | M | T | F | ? | ? | F | F | L | F | T | F | ? | M |
| | H2 | T | T | F | F | ? | F | F | ? | F | ? | T | ? | F | M | F | T | F | T | F | F | H | ? | ? | F | ? | F | F | L | F | T | F | F | H |
| Consen. | L | 6 | 6 | -6 | -3 | 2 | -6 | -5 | 6 | -6 | 1 | 3 | -3 | -4 | MH | -6 | 3 | -2 | 6 | -1 | -6 | MH | 3 | -6 | -3 | 1 | -3 | -4 | M | -1 | 6 | -6 | 4 | MH |
| | H | 2 | 2 | -2 | -2 | 0 | -2 | -2 | 0 | -2 | 0 | 1 | -1 | -2 | LM | -2 | 2 | -2 | 2 | -2 | -2 | MH | 1 | -1 | -1 | 0 | -2 | -2 | L | -2 | 2 | -2 | -1 | MH |
| | L+H | 8 | 8 | -8 | -5 | 2 | -8 | -7 | 6 | -8 | 1 | 4 | -4 | -6 | M | -8 | 5 | -4 | 8 | -3 | -8 | MH | 4 | -7 | -4 | 1 | -5 | -6 | LM | -3 | 8 | -8 | 3 | MH |

**Figure 1(i):** Classification results for trials 1-5

**Key:** LLM reviewers L1-6 are defined in Table 4. H1 and H2 are human reviewers. Consen. L and H indicates LLM and Human consensus. Confidence levels L, M and H indicate Low, Medium and High.

| NCT Number | Classifier | AI Task Category | | | | | | | | | | | | | | Human-AI Relationship | | | | | | | Human-AI Interaction | | | | | | | Humans Interacting with AI | | | | |
|---|---|---|---|---|---|---|---|---|---|---|---|---|---|---|---|---|---|---|---|---|---|---|---|---|---|---|---|---|---|---|---|---|---|---|
| | | Classify | Detect/assess | Diagnose | Educate | Explain/interpret | Monitor/respond | Natural language communication | Personalise | Predict/prognose | Recommend/advise | Score/rank | Summarise/synthesise | Treat | AI Task Category Confidence | AI as a collaborator | AI as a decision-support resource | AI as a therapeutic component | AI as a tool | AI as an educator/communicator | AI as proactive or autonomous | Human-AI Relationship Confidence | AI first: human-in-the-loop | Human-out-of-the-loop | AI follows: human-in-the-loop | Human-on-the loop | Multiple humans and/or AIs | Iterative interaction | Human-AI Interaction Confidence | Caregiver-AI Interaction | Health Professional-AI Interaction | Other Human-AI Interaction | Patient AI-Interaction | Humans Interacting with AI Confidence |
| NCT04679181 | L1 | T | F | F | F | F | T | T | F | F | T | F | F | F | H | F | T | F | T | T | F | H | T | F | ? | T | F | T | H | T | T | F | F | H |
| | L2 | T | T | T | F | F | F | F | F | F | T | F | F | F | H | F | T | F | T | F | F | M | T | F | F | F | F | F | M | T | T | F | F | H |
| | L3 | T | ? | F | F | F | F | F | F | F | T | F | F | F | M | F | T | ? | T | F | F | M | T | F | ? | T | ? | F | M | ? | T | F | ? | M |
| | L4 | T | F | T | F | ? | F | F | F | F | T | F | F | F | M | F | T | F | T | F | F | M | T | F | F | ? | T | F | M | T | T | F | F | M |
| | L5 | T | ? | F | F | F | F | F | F | F | T | F | F | F | M | ? | T | F | T | ? | F | M | T | F | ? | F | T | F | M | T | T | F | ? | M |
| | L6 | ? | ? | F | F | F | F | ? | ? | F | F | ? | ? | F | L | F | ? | ? | T | F | F | L | ? | F | F | ? | T | ? | L | F | T | F | ? | M |
| | H1 | T | ? | ? | F | ? | F | F | F | F | T | F | F | F | M | ? | T | F | T | F | F | M | T | F | ? | ? | ? | F | L | ? | T | F | F | M |
| | H2 | F | F | ? | F | F | F | ? | F | F | F | F | ? | F | L | ? | ? | F | T | ? | F | L | ? | F | F | ? | ? | F | L | ? | ? | F | ? | L |
| Consen. | L | 5 | -1 | -2 | -6 | -5 | -4 | -3 | -5 | -6 | 4 | -5 | -5 | -6 | M | -5 | 5 | -4 | 6 | -3 | -6 | M | 5 | -6 | -3 | 0 | 1 | -3 | M | 3 | 6 | -6 | -3 | MH |
| | H | 0 | -1 | 0 | -2 | -1 | -2 | -1 | -2 | -2 | 0 | -2 | -1 | -2 | LM | 0 | 1 | -2 | 2 | -1 | -2 | LM | 1 | -2 | -1 | 0 | 0 | -2 | L | 0 | 1 | -2 | -1 | LM |
| | L+H | 5 | -2 | -2 | -8 | -6 | -6 | -4 | -7 | -8 | 4 | -7 | -6 | -8 | M | -5 | 6 | -6 | 8 | -4 | -8 | M | 6 | -8 | -4 | 0 | 1 | -5 | LM | 3 | 7 | -8 | -4 | M |
| NCT04763980 | L1 | T | F | F | T | T | F | T | T | F | F | F | F | F | H | F | T | F | T | T | F | H | T | F | ? | T | F | T | M | F | T | T | F | H |
| | L2 | F | F | F | T | ? | F | F | ? | F | ? | F | F | F | H | F | F | F | T | T | F | H | T | F | F | T | F | T | M | F | T | F | T | H |
| | L3 | ? | ? | F | T | ? | F | T | ? | F | ? | ? | ? | F | M | ? | T | F | T | T | F | M | T | F | ? | T | ? | F | M | ? | T | F | T | M |
| | L4 | F | F | F | T | T | F | T | T | F | F | F | F | F | M | T | T | F | T | T | F | M | T | F | F | F | F | T | M | F | T | F | F | M |
| | L5 | T | F | F | T | ? | F | T | F | F | F | F | F | F | M | T | T | ? | F | T | F | M | T | F | F | ? | T | T | H | ? | T | F | T | H |
| | L6 | F | F | F | T | T | F | T | ? | F | ? | F | ? | F | M | ? | ? | F | T | T | F | M | T | F | F | ? | T | T | M | F | ? | T | ? | M |
| | H1 | ? | ? | F | T | F | F | ? | ? | F | ? | F | F | F | L | ? | ? | F | T | ? | F | L | ? | F | F | ? | F | ? | L | F | T | ? | ? | L |
| | H2 | F | F | F | T | ? | F | T | T | F | F | F | F | F | M | F | F | F | F | T | F | M | F | F | F | F | F | T | H | F | ? | F | T | M |
| Consen. | L | -1 | -5 | -6 | 6 | 3 | -6 | 4 | 1 | -6 | -3 | -5 | -4 | -6 | MH | 0 | 3 | -5 | 4 | 6 | -6 | MH | 6 | -6 | -4 | 2 | -1 | 4 | M | -4 | 5 | -2 | 1 | MH |
| | H | -1 | -1 | -2 | 2 | -1 | -2 | 1 | 1 | -2 | -1 | -2 | -2 | -2 | LM | -1 | -1 | -2 | 0 | 1 | -2 | LM | -1 | -2 | -2 | -1 | -2 | 1 | M | -2 | 1 | -1 | 1 | LM |
| | L+H | -2 | -6 | -8 | 8 | 2 | -8 | 5 | 2 | -8 | -4 | -7 | -6 | -8 | M | -1 | 2 | -7 | 4 | 7 | -8 | M | 5 | -8 | -6 | 1 | -3 | 5 | M | -6 | 6 | -3 | 2 | MH |
| NCT05308303 | L1 | T | T | F | F | T | F | T | F | F | F | F | T | F | H | F | T | F | T | T | F | H | T | F | ? | T | F | T | H | T | T | F | F | H |
| | L2 | T | T | F | F | F | F | F | F | F | F | F | F | F | H | F | F | F | T | F | F | H | T | F | T | T | F | F | H | F | T | F | F | H |
| | L3 | T | ? | F | F | F | F | F | F | T | ? | F | T | F | M | F | ? | F | T | F | T | M | ? | T | F | ? | F | F | M | ? | T | F | F | M |
| | L4 | T | ? | F | F | F | F | T | F | F | F | F | ? | F | M | F | T | F | T | ? | F | M | T | F | ? | T | ? | F | M | F | T | F | F | M |
| | L5 | T | ? | F | F | F | F | F | F | F | ? | F | F | F | M | F | T | F | T | F | F | H | T | F | ? | F | F | F | H | T | F | F | F | M |
| | L6 | T | T | F | F | F | F | F | F | F | F | F | T | F | H | F | ? | F | T | F | F | H | T | F | F | T | ? | F | M | F | ? | T | F | M |
| | H1 | T | ? | F | F | F | F | F | F | F | F | F | ? | F | M | F | ? | F | T | F | ? | M | F | ? | ? | T | F | F | M | F | T | F | F | H |
| | H2 | T | F | F | F | T | F | T | F | F | F | F | T | F | M | F | F | F | T | F | F | M | ? | ? | F | ? | F | F | L | F | T | F | F | H |
| Consen. | L | 6 | 3 | -6 | -6 | -4 | -6 | -2 | -6 | -4 | -4 | -6 | 1 | -6 | MH | -6 | 2 | -6 | 6 | -3 | -4 | MH | 5 | -4 | -1 | 3 | -4 | -4 | MH | -1 | 3 | -4 | -6 | MH |
| | H | 2 | -1 | -2 | -2 | 0 | -2 | 0 | -2 | -2 | -2 | -2 | 1 | -2 | M | -2 | -1 | -2 | 2 | -2 | -1 | M | -1 | 0 | -1 | 1 | -2 | -2 | LM | -2 | 2 | -2 | -2 | H |
| | L+H | 8 | 2 | -8 | -8 | -4 | -8 | -2 | -8 | -6 | -6 | -8 | 2 | -8 | MH | -8 | 1 | -8 | 8 | -5 | -5 | MH | 4 | -4 | -2 | 4 | -6 | -6 | MH | -3 | 5 | -6 | -8 | MH |
| NCT05384002 | L1 | T | T | T | F | T | F | F | F | T | F | F | F | F | H | F | T | F | T | F | F | H | T | F | ? | T | F | F | M | T | T | F | F | H |
| | L2 | T | T | T | F | T | F | F | F | T | F | T | F | F | H | F | T | F | T | T | F | M | T | F | T | T | F | F | H | F | T | F | F | H |
| | L3 | T | T | ? | F | T | F | F | F | T | ? | F | F | F | H | F | T | F | T | F | ? | M | T | F | ? | T | ? | F | M | ? | T | ? | ? | L |
| | L4 | T | T | T | F | T | F | F | F | T | F | F | F | F | M | T | T | F | T | F | F | M | T | F | F | T | F | F | M | F | T | F | F | M |
| | L5 | T | ? | F | F | T | F | F | F | T | F | T | F | F | M | ? | T | F | T | F | F | M | T | F | ? | T | T | F | M | T | T | F | F | H |
| | L6 | T | T | T | F | T | F | F | T | T | ? | ? | F | F | H | F | T | F | T | F | F | M | ? | F | F | ? | F | F | L | F | ? | ? | F | L |
| | H1 | T | T | F | F | F | F | F | F | T | ? | ? | F | F | M | F | T | F | T | F | F | H | ? | F | ? | T | F | F | M | F | T | F | F | H |
| | H2 | T | T | T | F | F | F | F | F | T | F | F | ? | F | H | F | T | F | T | F | F | H | ? | ? | F | ? | F | F | L | F | T | F | ? | M |
| Consen. | L | 6 | 5 | 3 | -6 | 6 | -6 | -6 | -4 | 6 | -4 | -1 | -6 | -6 | MH | -3 | 6 | -6 | 6 | -4 | -5 | M | 5 | -6 | -1 | 5 | -3 | -6 | M | -1 | 5 | -4 | -5 | M |
| | H | 2 | 2 | 0 | -2 | -2 | -2 | -2 | -2 | 2 | -1 | -1 | -1 | -2 | MH | -2 | 2 | -2 | 2 | -2 | -2 | H | 0 | -1 | -1 | 1 | -2 | -2 | LM | -2 | 2 | -2 | -1 | MH |
| | L+H | 8 | 7 | 3 | -8 | 4 | -8 | -8 | -6 | 8 | -5 | -2 | -7 | -8 | MH | -5 | 8 | -8 | 8 | -6 | -7 | MH | 5 | -7 | -2 | 6 | -5 | -8 | M | -3 | 7 | -6 | -6 | MH |
| NCT06070415 | L1 | T | F | F | T | T | F | T | T | F | T | F | F | F | H | F | T | F | T | T | F | H | T | F | ? | T | F | T | H | F | T | T | T | H |
| | L2 | F | F | F | T | T | F | F | F | F | F | F | F | F | H | F | F | F | T | T | F | H | T | F | F | F | F | ? | M | F | F | F | T | H |
| | L3 | ? | F | F | T | ? | F | T | ? | F | ? | F | F | F | M | ? | F | F | T | T | F | M | T | F | ? | T | F | T | M | ? | ? | F | T | M |
| | L4 | F | F | F | T | F | F | T | ? | F | F | F | F | F | M | ? | F | F | T | T | F | M | T | F | F | ? | F | T | M | F | ? | F | T | M |
| | L5 | T | ? | F | ? | F | F | T | F | F | T | F | F | F | M | T | ? | F | T | T | F | M | T | F | ? | ? | T | T | M | F | ? | F | T | M |
| | L6 | F | F | F | T | T | F | T | F | F | T | F | F | T | H | ? | F | T | T | T | ? | M | ? | ? | F | F | F | ? | M | F | ? | F | T | H |
| | H1 | ? | F | F | T | T | F | T | ? | F | ? | F | F | F | H | F | F | ? | T | T | F | M | T | F | ? | ? | F | T | M | F | F | F | T | H |
| | H2 | F | F | F | T | F | F | F | F | F | F | F | F | F | M | F | F | T | F | F | F | M | F | ? | F | F | F | ? | L | F | F | F | T | H |
| Consen. | L | -1 | -5 | -6 | 5 | 1 | -6 | 4 | -2 | -6 | 1 | -6 | -6 | -4 | MH | -1 | -3 | -4 | 6 | 6 | -5 | MH | 5 | -5 | -3 | 0 | -4 | 4 | M | -5 | 0 | -4 | 6 | MH |
| | H | -1 | -2 | -2 | 2 | 0 | -2 | 0 | -1 | -2 | -1 | -2 | -2 | -2 | MH | -2 | -2 | 1 | 0 | 0 | -2 | M | 0 | -1 | -1 | -1 | -2 | 1 | LM | -2 | -2 | -2 | 2 | H |
| | L+H | -2 | -7 | -8 | 7 | 1 | -8 | 4 | -3 | -8 | 0 | -8 | -8 | -6 | MH | -3 | -5 | -3 | 6 | 6 | -7 | MH | 5 | -6 | -4 | -1 | -6 | 5 | M | -7 | -2 | -6 | 8 | MH |

**Figure 1(ii):** Classification results for trials 6-10

**Key:** LLM reviewers L1-6 are defined in Table 4. H1 and H2 are human reviewers. Consen. L and H indicates LLM and Human consensus. Confidence levels L, M and H indicate Low, Medium and High.

| | | AI Task Category | | | | | | | | | | | | | | Human-AI Relationship | | | | | | | Human-AI Interaction | | | | | | | Humans Interacting with AI | | | | |
|---|---|---|---|---|---|---|---|---|---|---|---|---|---|---|---|---|---|---|---|---|---|---|---|---|---|---|---|---|---|---|---|---|---|---|
| NCT Number | Classifier | Classify | Detect/assess | Diagnose | Educate | Explain/interpret | Monitor/respond | Natural language communication | Personalise | Predict/prognose | Recommend/advise | Score/rank | Summarise/synthesise | Treat | AI Task Category Confidence | AI as a collaborator | AI as a decision-support resource | AI as a therapeutic component | AI as a tool | AI as an educator/communicator | AI as proactive or autonomous | Human-AI Relationship Confidence | AI first: human-in-the-loop | Human-out-of-the-loop | AI follows: human-in-the-loop | Human-on-the loop | Multiple humans and/or AIs | Iterative interaction | Human-AI Interaction Confidence | Caregiver-AI Interaction | Health Professional-AI Interaction | Other Human-AI Interaction | Patient AI-Interaction | Humans Interacting with AI Confidence |
| NCT06186557 | L1 | T | T | F | F | T | F | T | F | F | F | F | F | F | H | F | T | F | T | T | F | H | T | F | ? | T | F | F | M | T | T | F | F | H |
| | L2 | T | T | F | F | F | F | F | F | F | F | F | F | F | H | F | F | F | T | F | F | H | T | F | F | F | F | F | H | F | T | F | F | H |
| | L3 | T | T | F | F | F | F | F | F | F | F | F | F | F | H | F | T | F | T | F | ? | M | T | F | F | ? | F | F | M | F | T | F | F | H |
| | L4 | T | T | F | F | F | F | F | F | F | F | T | F | F | M | F | T | F | T | F | F | M | T | F | F | T | F | F | M | F | T | F | F | H |
| | L5 | T | T | F | F | ? | F | F | F | F | ? | F | F | F | H | F | T | F | T | F | F | H | T | F | ? | F | F | F | M | ? | T | F | F | M |
| | L6 | T | T | F | F | F | F | F | F | F | F | F | F | F | H | F | F | F | T | F | F | H | ? | F | ? | T | T | F | M | F | ? | T | F | M |
| | H1 | T | T | F | F | ? | F | F | F | F | F | F | F | F | M | F | T | F | T | F | F | H | T | ? | ? | T | F | F | M | F | T | F | F | H |
| | H2 | T | T | T | F | T | ? | F | F | T | ? | ? | F | F | M | F | T | F | T | F | ? | M | ? | ? | ? | ? | F | F | L | F | T | F | ? | M |
| Consen. | L | 6 | 6 | -6 | -6 | -3 | -6 | -4 | -6 | -6 | -5 | -4 | -6 | -6 | H | -6 | 2 | -6 | 6 | -4 | -5 | MH | 5 | -6 | -3 | 1 | -4 | -6 | M | -3 | 5 | -4 | -6 | MH |
| | H | 2 | 2 | 0 | -2 | 1 | -1 | -2 | -2 | 0 | -1 | -1 | -2 | -2 | M | -2 | 2 | -2 | 2 | -2 | -1 | MH | 1 | 0 | 0 | 1 | -2 | -2 | LM | -2 | 2 | -2 | -1 | MH |
| | L+H | 8 | 8 | -6 | -8 | -2 | -7 | -6 | -8 | -6 | -6 | -5 | -8 | -8 | MH | -8 | 4 | -8 | 8 | -6 | -6 | MH | 6 | -6 | -3 | 2 | -6 | -8 | M | -5 | 7 | -6 | -7 | MH |
| NCT06265909 | L1 | T | T | F | F | T | T | T | F | T | F | F | F | F | H | F | T | F | T | T | F | H | T | F | T | T | ? | T | H | F | T | F | T | H |
| | L2 | F | T | F | F | F | F | F | T | T | T | F | F | F | H | F | T | F | T | F | F | H | ? | F | F | F | F | F | M | F | F | F | T | H |
| | L3 | T | ? | F | F | F | F | F | T | T | ? | F | F | F | M | ? | T | F | T | F | ? | M | T | ? | F | T | F | F | M | F | F | ? | T | H |
| | L4 | T | T | F | F | F | T | F | T | T | F | F | F | F | H | F | T | F | T | F | F | H | T | F | F | T | ? | ? | M | F | F | F | T | H |
| | L5 | T | T | F | ? | ? | F | F | T | T | ? | F | F | F | M | ? | T | F | T | ? | ? | M | T | ? | F | T | F | F | M | F | T | F | T | H |
| | L6 | F | T | F | F | F | F | F | T | T | T | T | F | ? | H | F | T | ? | T | F | ? | M | ? | ? | ? | F | F | ? | L | F | F | F | T | H |
| | H1 | T | T | F | F | ? | F | F | ? | ? | T | F | F | ? | M | F | ? | ? | T | F | ? | L | ? | ? | F | ? | F | F | L | F | T | ? | T | M |
| | H2 | T | T | F | F | F | F | F | T | T | T | T | F | T | M | F | T | T | T | F | T | H | F | T | F | F | F | F | M | F | F | F | T | H |
| Consen. | L | 2 | 5 | -6 | -5 | -3 | -2 | -4 | 4 | 6 | 0 | -4 | -6 | -5 | MH | -4 | 6 | -5 | 6 | -3 | -3 | MH | 4 | -3 | -3 | 2 | -4 | -2 | M | -6 | -2 | -5 | 6 | H |
| | H | 2 | 2 | -2 | -2 | -1 | -2 | -2 | 1 | 1 | 2 | 0 | -2 | 1 | M | -2 | 1 | 1 | 2 | -2 | 1 | M | -1 | 1 | -2 | -1 | -2 | -2 | LM | -2 | 0 | -1 | 2 | MH |
| | L+H | 4 | 7 | -8 | -7 | -4 | -4 | -6 | 5 | 7 | 2 | -4 | -8 | -4 | MH | -6 | 7 | -4 | 8 | -5 | -2 | MH | 3 | -2 | -5 | 1 | -6 | -4 | M | -8 | -2 | -6 | 8 | H |
| NCT06423040 | L1 | T | F | F | T | F | T | T | F | F | F | F | F | F | H | F | T | F | T | T | F | H | T | F | ? | T | F | T | M | F | F | T | T | H |
| | L2 | F | F | F | T | F | F | F | F | F | T | F | F | F | H | F | F | F | T | T | F | H | T | F | F | F | F | T | H | F | F | F | T | H |
| | L3 | ? | ? | F | T | ? | ? | T | ? | F | ? | ? | ? | F | M | F | ? | F | T | T | ? | M | T | ? | F | ? | ? | T | M | F | F | ? | T | M |
| | L4 | F | F | F | T | F | T | T | ? | F | T | F | F | F | M | F | F | F | T | T | T | M | F | T | F | F | F | T | M | F | F | F | T | M |
| | L5 | T | ? | F | T | ? | F | T | ? | F | T | F | F | F | M | F | ? | F | T | T | F | M | T | F | ? | ? | ? | T | M | F | F | F | T | H |
| | L6 | F | F | F | T | T | ? | T | F | F | T | F | F | F | H | T | F | T | T | T | ? | H | ? | ? | F | ? | F | T | M | F | F | F | T | H |
| | H1 | F | F | F | T | ? | F | T | ? | F | ? | F | F | F | M | ? | F | T | T | T | ? | M | ? | ? | F | F | F | T | M | F | F | F | T | H |
| | H2 | F | F | F | T | F | ? | T | ? | F | ? | F | F | F | M | F | F | T | F | T | T | M | F | F | F | F | F | ? | L | F | F | F | T | H |
| Consen. | L | -1 | -4 | -6 | 6 | -2 | 0 | 4 | -3 | -6 | 3 | -5 | -5 | -6 | MH | -4 | -2 | -4 | 6 | 6 | -2 | MH | 3 | -2 | -4 | -1 | -4 | 6 | M | -6 | -6 | -3 | 6 | MH |
| | H | -2 | -2 | -2 | 2 | -1 | -1 | 2 | 0 | -2 | 0 | -2 | -2 | -2 | M | -1 | -2 | 2 | 0 | 2 | 1 | M | -1 | -1 | -2 | -2 | -2 | 1 | LM | -2 | -2 | -2 | 2 | H |
| | L+H | -3 | -6 | -8 | 8 | -3 | -1 | 6 | -3 | -8 | 3 | -7 | -7 | -8 | MH | -5 | -4 | -2 | 6 | 8 | -1 | MH | 2 | -3 | -6 | -3 | -6 | 7 | M | -8 | -8 | -5 | 8 | H |
| NCT06614530 | L1 | T | F | F | F | F | ? | F | T | F | F | F | F | F | M | F | T | T | T | F | F | H | F | F | ? | T | F | F | M | F | T | F | F | H |
| | L2 | F | T | F | F | F | T | F | T | F | F | F | F | F | M | F | F | F | T | F | T | M | F | T | F | F | F | F | M | F | F | F | T | H |
| | L3 | ? | T | F | F | F | T | F | ? | F | ? | F | F | T | M | ? | T | T | T | F | F | M | ? | F | ? | T | F | F | M | ? | ? | F | T | M |
| | L4 | F | T | F | F | F | T | F | T | F | F | F | F | F | M | F | F | T | T | F | T | M | T | T | F | F | F | F | M | F | F | F | T | H |
| | L5 | T | T | F | F | F | T | F | T | F | F | F | F | F | M | F | T | ? | T | F | ? | M | T | F | F | ? | F | F | M | F | T | F | T | H |
| | L6 | ? | T | F | F | F | T | F | T | F | F | F | F | T | H | F | F | T | T | F | T | H | F | T | F | ? | F | ? | H | F | F | F | T | H |
| | H1 | F | T | F | F | F | ? | F | T | F | F | F | F | F | H | F | F | T | T | F | T | M | F | T | F | F | F | F | M | F | F | F | T | M |
| | H2 | F | F | F | F | F | T | F | T | F | F | F | F | T | H | F | F | F | T | F | T | H | F | T | F | F | F | F | H | F | F | F | T | H |
| Consen. | L | 0 | 4 | -6 | -6 | -6 | 5 | -6 | 5 | -6 | -5 | -6 | -6 | -2 | M | -5 | 0 | 3 | 6 | -6 | 1 | MH | -1 | 0 | -4 | 0 | -6 | -5 | M | -5 | -1 | -6 | 4 | H |
| | H | -2 | 0 | -2 | -2 | -2 | 1 | -2 | 2 | -2 | -2 | -2 | -2 | 0 | H | -2 | -2 | 0 | 2 | -2 | 2 | MH | -2 | 2 | -2 | -2 | -2 | -2 | MH | -2 | -2 | -2 | 2 | MH |
| | L+H | -2 | 4 | -8 | -8 | -8 | 6 | -8 | 7 | -8 | -7 | -8 | -8 | -2 | MH | -7 | -2 | 3 | 8 | -8 | 3 | MH | -3 | 2 | -6 | -2 | -8 | -7 | MH | -7 | -3 | -8 | 6 | H |
| NCT06664112 | L1 | T | T | F | F | T | T | F | F | T | T | F | T | F | H | F | T | F | T | F | F | M | T | F | T | T | F | T | H | T | T | F | F | H |
| | L2 | T | T | F | F | T | F | F | F | T | F | F | F | F | H | F | T | F | T | T | F | H | T | F | T | T | F | F | H | F | T | F | F | H |
| | L3 | T | T | F | F | T | F | F | F | T | T | F | F | F | H | ? | T | F | T | F | F | H | T | F | ? | T | F | F | H | ? | T | F | F | H |
| | L4 | T | T | F | F | T | F | F | ? | T | T | T | F | F | M | T | T | F | T | ? | F | M | T | F | ? | T | F | F | M | F | T | F | F | M |
| | L5 | T | ? | F | F | T | F | F | F | ? | T | F | F | F | M | ? | T | F | T | F | F | M | T | F | ? | F | F | F | L | T | T | F | F | H |
| | L6 | T | T | ? | F | T | ? | F | T | T | T | T | T | F | H | F | T | F | T | ? | F | H | T | F | F | T | T | ? | H | F | T | F | ? | H |
| | H1 | T | T | F | F | ? | F | F | F | T | F | ? | F | F | M | F | T | F | T | F | F | H | T | F | ? | ? | F | F | M | F | T | ? | F | M |
| | H2 | T | T | F | F | F | F | F | F | T | F | T | F | F | H | F | T | F | T | F | F | H | T | F | ? | F | F | F | M | F | T | F | ? | M |
| Consen. | L | 6 | 5 | -5 | -6 | 6 | -3 | -6 | -3 | 5 | 4 | -2 | -2 | -6 | MH | -2 | 6 | -6 | 6 | -2 | -6 | MH | 6 | -6 | 1 | 4 | -4 | -3 | MH | -1 | 6 | -6 | -5 | H |
| | H | 2 | 2 | -2 | -2 | -1 | -2 | -2 | -2 | 2 | -2 | 1 | -2 | -2 | MH | -2 | 2 | -2 | 2 | -2 | -2 | H | 2 | -2 | 0 | -1 | -2 | -2 | M | -2 | 2 | -1 | -1 | M |
| | L+H | 8 | 7 | -7 | -8 | 5 | -5 | -8 | -5 | 7 | 2 | -1 | -4 | -8 | MH | -4 | 8 | -8 | 8 | -4 | -8 | MH | 8 | -8 | 1 | 3 | -6 | -5 | MH | -3 | 8 | -7 | -6 | MH |

**Figure 1(iii):** Classification results for trials 11-15

**Key:** LLM reviewers L1-6 are defined in Table 4. H1 and H2 are human reviewers. Consen. L and H indicates LLM and Human consensus. Confidence levels L, M and H indicate Low, Medium and High.

**Table 6:** Categories most frequently classified as True, False and Uncertain across all eight classifiers.

| True | | |
|---|---|---|
| **Dimension** | **Category** | **No. of classifications (/120)** |
| Human-AI Relationship | AI as a tool | 114 |
| AI Task Category | Classify | 82 |
| Human-AI Interaction | AI first; human-in-the-loop | 76 |
| Humans Interacting with AI | Health Professional-AI Interaction | 75 |
| Human-AI Relationship | AI as a decision-support resource | 73 |
| **False** | | |
| **Dimension** | **Category** | **No. of classifications (/120)** |
| AI Task Category | Treat | 108 |
| AI Task Category | Diagnose | 104 |
| Humans Interacting with AI | Other Human-AI Interaction | 103 |
| AI Task Category | Monitor/respond | 99 |
| AI Task Category | Summarise/synthesise | 98 |
| **Uncertain** | | |
| **Dimension** | **Category** | **No. of classifications (/120)** |
| Human-AI Interaction | AI follows; human-in-the-loop | 47 |
| Human-AI Interaction | Human-on the loop | 38 |
| AI Task Category | Detect/assess | 27 |
| Human-AI Interaction | AI first; human-in-the-loop | 26 |
| AI Task Category | Recommend/advise | 25 |

**Table 7:** Categories with the highest and lowest levels of agreement among the eight classifiers.

| Highest agreement | | |
|---|---|---|
| **Dimension** | **Category** | **Average agreement (/8)** |
| Human-AI Relationship | AI as a tool | 7.6 |
| AI Task Category | Predict/prognose | 7.6 |
| AI Task Category | Educate | 7.4 |
| AI Task Category | Treat | 7.2 |
| AI Task Category | Diagnose | 7.1 |
| **Lowest agreement** | | |
| **Dimension** | **Category** | **Average agreement (/8)** |
| Human-AI Interaction | Human-on the loop | 4.1 |
| Human-AI Interaction | AI follows; human-in-the-loop | 4.9 |
| AI Task Category | Recommend/advise | 5.3 |
| Human-AI Interaction | AI first; human-in-the-loop | 5.3 |
| AI Task Category | Explain/interpret | 5.5 |

## 7. Discussion

Prior work and iterative piloting of the dimensions and categories highlighted the ambiguous nature of many trial narratives and the subjective nature of the categorisation tasks. Indeed, the trial set was purposively sampled to include trials where categorisation was known to be difficult.

Several definitions were usefully clarified during the process. For example, patient-AI interaction was clarified as absent where AI operated only on patient-derived data and patients did not directly interact with the AI system or its outputs.

An example of ambiguity in a trial record was NCT00647413, which involved the use of an expert system to tailor a letter to parents who smoked, based on urinary cotinine (a measure of nicotine exposure) in their infant's diapers. Counselling also formed part of the intervention; however, it was not clear from the inspected fields whether the expert system contributed to the counselling. Given the trial's 2008 start date and the capabilities of expert systems at that time, it could be inferred that it did not provide the counselling. The trial record also did not indicate whether the letters were reviewed by a health professional before sent, leaving the nature of human oversight uncertain.

A further example of ambiguity regarding human oversight was NCT04289025, which involved the use of a GaitSmart medical device for gait assessment, the provision of a report to an intervention group, and six personalised exercises. Nurses ensured that participants could complete the exercises before leaving, and unsuitable exercises were removed. However, it was unclear whether the AI-generated report was reviewed and whether the personalised exercises were recommended by the nurses or generated by the AI.

Some of most challenging trials to categorise in terms of HAII were, in fact, those focussed on AI development and on HAII itself. For example, NCT03780582 involved the evaluation of diagnostic AI for lung cancer across three different HAII configurations, in which radiologists were variously presented with AI predictions and regions of interest, and the outcomes were compared. To address these trials, it was agreed that an 'AI in development/training/optimisation' category was required.

These examples reflect substantial ambiguity in some trials, particularly where records refer to the use of AI but provide little or no detail about AI methods, interaction sequencing or human oversight.

It was also agreed that measures of AI performance, such as sensitivity, specificity, precision and recall, should not be classified as scoring/ranking tasks. Instead, the AI task should reflect the clinical or health-related function performed by the system.

The findings indicate a potential role for AI tools in supporting the completion of trial records by encouraging the provision of additional detail and highlighting ambiguous descriptions of AI methods and human-AI interactions.

## 8. Conclusions and Further Work

This study developed and preliminarily evaluated a multidimensional framework for categorising human-AI interactions in clinical trials. The framework extends previous categorisation approaches by incorporating AI tasks, human-AI relationships, interaction configurations and the interacting human groups. The research questions were addressed through the definitions of HAII, the development and application of the framework, and the presentation of preliminary comparison of human and LLM classifications. RQ1 was addressed through a definition of HAII based on human perception of AI outputs and AI detection of human inputs. RQ2 was addressed through the development of a multidimensional framework encompassing AI tasks, human-AI relationships, interaction configurations and interacting human groups. RQ3 was addressed through comparison of human and LLM classifications, highlighting both agreement and areas of ambiguity.

Across the selected trials, AI was most frequently categorised as a tool, AI-first human-in-the-loop interaction was the most frequent interaction configuration, and health professionals were the human group most frequently identified as interacting with AI.

Application of the framework highlighted substantial ambiguity in the reporting of AI functionality, interaction sequencing and human oversight within clinical-trial records. The findings also suggest that LLMs may have value as classification aids, although human judgement remains important where trial records are incomplete or ambiguous.

Further work will refine the framework, clarify category definitions and decision rules, and evaluate its application across a larger sample of clinical trials. This refinement will include an AI in development/training/optimisation category. Such studies differ from evaluations of deployed AI because their primary purpose is often the development, refinement or validation of the AI system itself rather than assessment of routine human-AI interaction.

The revised framework will initially be applied to the fuller set of 100 trials in the previously reported dataset before being extended to the wider dataset. In conclusion, the findings reported here indicate a need not only for reporting guidance, but also for practical support for the consistent description and categorisation of human-AI interactions in clinical trials.

### Acknowledgments

Authors gratefully acknowledge support of the Digital Society Institute at Keele University, UK, that underpins efforts towards the publication of work reported in this paper. For the purposes of open access, the authors have applied a Creative Commons Attribution (CC-BY) license to any Accepted Author Manuscript version arising from this submission. In addition to extensive use of LLMs as part of the empirical investigation, the authors also acknowledge the limited use of enterprise Microsoft 365 Copilot GPT 5.6 to support preliminary analysis of data and preparation of this manuscript.

## References

[1] C. Jacob, N. Brasier, E. Laurenzi, S. Heuss, S.-G. Mougiakakou, A. Cöltekin, M. K. Peter, AI for IMPACTS framework for evaluating the long-term real-world impacts of AI-powered clinician tools: systematic review and narrative synthesis, Journal of Medical Internet Research 27, 2025. e67485. doi: 10.2196/67485.

[2] S. Joshi, I. Urteaga, W. A. C. van Amsterdam, G. Hripcsak, P. Elias, B. Recht, N. Elhadad, J. Fackler, M. P. Sendak, J. Wiens, AI as an intervention: improving clinical outcomes relies on a causal approach to AI development and validation, Journal of the American Medical Informatics Association 32, 2025. pp. 589–594. doi: 10.1093/jamia/ocae301.

[3] ClinicalTrials.gov (2026) *Trends and charts on registered studies*. Bethesda, MD: U.S. National Library of Medicine. Accessed 17 August 2026.

[4] S. Woolley, T. Collins, K. Khattak, I. Chernomorets, A. Arevalo, C. Richardson, Trends in AI and Human-AI Interaction in Clinical Trials—A Hybrid Human-AI Exploration, in: Workshop on Health, Wellbeing and Human AI-Interaction, Hybrid-Human AI Conference, 2026. arXiv: 2605.29096.

[5] S. Cruz Rivera, X. Liu, A.-W. Chan, A. K. Denniston, M. J. Calvert, SPIRIT-AI and CONSORT-AI Working Group, Guidelines for clinical trial protocols for interventions involving artificial intelligence: the SPIRIT-AI extension, Nature Medicine 26, 2020, pp. 1351–1363. doi: 10.1038/s41591-020-1037-7.

[6] X. Liu, S. Cruz Rivera, D. Moher, M. J. Calvert, A. K. Denniston, SPIRIT-AI and CONSORT-AI Working Group, Reporting guidelines for clinical trial reports for interventions involving artificial intelligence: the CONSORT-AI extension, BMJ 370, 2020, m3164. doi: 10.1136/bmj.m3164.

[7] N. Pattathil, J. Z. L. Zhao, O. Sam-Oyerinde, T. Felfeli, Adherence of randomised controlled trials using artificial intelligence in ophthalmology to CONSORT-AI guidelines: a systematic review and critical appraisal, BMJ Health & Care Informatics 30, 2023, e100757. doi: 10.1136/bmjhci-2023-100757.

[8] D. Chen, K. Arnold, R. Sukhdeo, J. F. Alla, S. Raman, Concordance with CONSORT-AI guidelines in reporting of randomised controlled trials investigating artificial intelligence in oncology: a systematic review, BMJ Oncology 4, 2025, e000733. doi: 10.1136/bmjonc-2025-000733.

[9] T. Collins, S. I. Woolley, S. Oniani, I. M. Pires, N. M. Garcia, S. J. Ledger, A. Pandyan, Version reporting and assessment approaches for new and updated activity and heart rate monitors, Sensors 19, 2019, 1705. doi: 10.3390/s19071705.

[10] K. A. Khattak, S. I. Woolley, T. Collins, Wearables, healthcare-computer interaction and the internet of obscure medical things, in: Proceedings of the 37th International BCS Human-Computer Interaction Conference, BCS Learning and Development, Swindon, 2024, pp. 225–229. doi: 10.14236/ewic/BCSHCI2024.22.

[11] T. Y. T. Lam, M. F. K. Cheung, Y. L. Munro, K. M. Lim, D. Shung, J. J. Y. Sung, Randomized Controlled Trials of Artificial Intelligence in Clinical Practice: Systematic Review. J Med Internet Res. 2022 Aug 25;24(8):e37188. doi: 10.2196/37188.

[12] Q. Wang, Y. Yang, L. Qiu, Y. Xu, R. Sun, P. Xue and Y. Qiao, A quantitative analysis of global AI medical studies: gaps in randomized controlled trials. *npj Digit. Med.* **9**, 403 (2026). doi: 10.1038/s41746-026-02698-z

[13] D. Plana, D. L. Shung, A. A. Grimshaw, A. Saraf, J. J. Y. Sung and B. H. Kann (2022) 'Randomized clinical trials of machine learning interventions in health care: a systematic review', *JAMA Network Open*, 5(9), e2233946. Available at: doi: 10.1001/jamanetworkopen.2022.33946.

[14] B. Brijnath, V. Welch, F. Habibzadeh, S. Dawson, T. Li, D. P. Richards, O. Chinembiri, D. Ghersi, A. M. Orkin, K. A. Cameron, D. Coase, R. M. Golub, M. Heuschkel, L. Jasicki, L. Leigh, R. Benn, G. Papadopoulos, N. Siegfried, N. Straiton, S. Treweek, A. Chan, H. Green, E. Owusu-Addo, P. Feldman, R. Muoio, Measurement of ethnicity in clinical trials: Delphi survey and consensus statement, BMJ 394 (2026) e409503. doi: 10.1136/bmj-2026-409503.

[15] D. B. Olawade, S. C. Fidelis, S. Marinze, E. Egbon, A. Osunmakinde, A. Osborne, Artificial intelligence in clinical trials: a comprehensive review of opportunities, challenges, and future directions, International Journal of Medical Informatics 206 (2026) 106141. doi: 10.1016/j.ijmedinf.2025.106141.

[16] P. Giannouris, T. Myridis, T. Passali, G. Tsoumakas, Plain language summarization of clinical trials, in: Proceedings of the Workshop on DeTermIt! Evaluating Text Difficulty in a Multilingual Context, Association for Computational Linguistics, 2024, pp. 45–59, Available at:

https://aclanthology.org/2024.determit-1.6/
[17] A. Srinivasan, J. Berkowitz, N. A. Friedrich, S. Kivelson, N. P. Tatonetti, Large language model analysis of reporting quality of randomized clinical trial articles: a systematic review, JAMA Network Open 8 (2025) e2529418. doi: 10.1001/jamanetworkopen.2025.29418.
[18] R. Han, J. N. Acosta, Z. Shakeri, J. P. A. Ioannidis, E. J. Topol, P. Rajpurkar, Randomised controlled trials evaluating artificial intelligence in clinical practice: a scoping review, The Lancet Digital Health 6 (2024) e367–e373. doi: 10.1016/S2589-7500(24)00047-5.
[19] C. Gomez, S. M. Cho, S. Ke, C.-M. Huang, M. Unberath, Human-AI collaboration is not very collaborative yet: a taxonomy of interaction patterns in AI-assisted decision making from a systematic review, Frontiers in Computer Science 6 (2025) 1521066. doi: 10.3389/fcomp.2024.1521066.
[20] Y. You, X. Li, A scoping review of human-AI collaboration patterns and task divisions in healthcare applications, npj Digital Medicine (2026). doi: 10.1038/s41746-026-02912-y.
[21] L. Madeyski, B. Kitchenham, M. Shepperd, LLM4SCREENLIT: Recommendations on assessing the performance of large language models for screening literature in systematic reviews, Information and Software Technology 198 (2026) 108204. doi: 10.1016/j.infsof.2026.108204.
[22] A. Dix, J. Finlay, G. D. Abowd, R. Beale, Human-computer interaction, 3rd ed., Pearson Education, Harlow, 2004.

# Appendix

## A. AI prompt (markdown format):

```
You are an expert systematic reviewer in clinical applications of AI with expertise in
Human-AI interaction.

This is a record of a clinical trial taken from the clinicaltrials.gov repository. Inspect
the record and classify the role of artificial intelligence in the clinical trial.

1. AI Task Category: Categorise the task that AI is performing, selecting all that apply
from:
* Detect/assess
* Classify
* Score/rank
* Diagnose (i.e. diagnostic labelling of patient condition, not just assessing a change or
progression of a pre-existing condition)
* Educate
* Explain/interpret
* Monitor/respond (i.e. continuous monitoring with responses when attention is required,
not monitoring by occasional assessment)
* Natural language communication
* Personalise
* Predict/prognose (i.e. predicting the future progression of a condition)
* Recommend/advise
* Summarise/synthesise
* Treat

2. Human-AI Relationship: Categorise the relationship between human/s and AI, selecting
all that apply from:
* AI as a tool (i.e. as a narrow function instrument)
* AI as an educator/communicator (i.e. advises human or communicates information to human
in natural language)
* AI as a decision-support resource (i.e. provides suggestions for actions or
interventions to aid decision making)
* AI as a collaborator (i.e. forms a two-way dialogue with human to collaboratively
achieve a goal)
* AI as a therapeutic/interventional component (i.e. forms an integral part of the
therapy, treatment or intervention)
* AI as proactive or autonomous (i.e. acts independently of human with no checking or
oversight)

3. Human-AI Interaction: Categorise the nature of the interaction between human/s and AI,
selecting all that apply from:
```

* AI outputs, human decides (AI first; human-in-the-loop)
* Human reviews/decides, then considers AI output (AI follows; human-in-the-loop)
* Iterative interaction (repeated exchanges, including chatbot interaction)
* Integrated interactions (multiple humans and/or multiple AIs)
* Humans oversee AI action (human-on-the loop)
* Autonomous AI action without human oversight (human-out-of-the-loop)

4. Humans Interacting with AI: Specify who the humans are that are interacting with AI, selecting all that apply from:
* Patient AI-Interaction
* Caregiver-AI Interaction
* Health Professional-AI Interaction
* Other Human-AI Interaction

5. Type of AI: Briefly state the type of AI that is being used (e.g. expert system, chatbot, neural network, etc.). Quote the type of AI directly from the trial record; do not assume or infer, but state that the type of AI is 'unknown' if it is not clearly stated.

6. Summary: Provide a short summary of the AI task. Describe what kind of AI is used. Describe how it interacts with humans. Explain who the humans are (e.g. clinicians, patients, caregivers, etc.). Explain the categorisations that you allocated in this summary.

For each of the four categorisations, provide an overall confidence rating of low, medium or high based on how clearly the trials record explained the use of AI and the nature of the human-AI interaction. The options for the individual categorisations are 'true' where the case is definitely true, 'false' where it is definitely not true, and 'uncertain' where there is insufficient information in the trials record to make a confident assessment. The 'true' and 'false' classifications should only be applied when the record makes the classification clear and unambiguous; do not make assumptions or inferences.

Use the trial description, intervention details, and other fields in the record to synthesise the summary and to determine the most appropriate categories. Produce a strict JSON output conforming to the schema specified.

While classifying, do not infer artificial intelligence use from vague digital terminology alone.